# Rethinking Image-based Table Recognition Using Weakly Supervised Methods

Nam Tuan Ly[a], Atsuhiro Takasu[b], Phuc Nguyen[c] and Hideaki Takeda[d]
*National Institute of Informatics (NII), Tokyo, Japan*
{namly, takasu, phucnt, takeda}@nii.ac.jp

Abstract: Most of the previous methods for table recognition rely on training datasets containing many richly annotated table images. Detailed table image annotation, e.g., cell or text bounding box annotation, however, is costly and often subjective. In this paper, we propose a weakly supervised model named WSTabNet for table recognition that relies only on HTML (or LaTeX) code-level annotations of table images. The proposed model consists of three main parts: an encoder for feature extraction, a structure decoder for generating table structure, and a cell decoder for predicting the content of each cell in the table. Our system is trained end-to-end by stochastic gradient descent algorithms, requiring only table images and their ground-truth HTML (or LaTeX) representations. To facilitate table recognition with deep learning, we create and release WikiTableSet, the largest publicly available image-based table recognition dataset built from Wikipedia. WikiTableSet contains nearly 4 million English table images, 590K Japanese table images, and 640k French table images with corresponding HTML representation and cell bounding boxes. The extensive experiments on WikiTableSet and two large-scale datasets: FinTabNet and PubTabNet demonstrate that the proposed weakly supervised model achieves better, or similar accuracies compared to the state-of-the-art models on all benchmark datasets.

## 1 INTRODUCTION

Table recognition has been receiving much attention from numerous researchers. It plays an important role in many document analysis systems, in which tabular data presented by PDF or document image often contains rich and essential information in a structured format. Table recognition aims to recognize the table structure and the content of each table cell from a table image, and to represent them in a machine-readable format, which can be HTML code (Jimeno Yepes et al., 2021; Li et al., 2019; Zhong et al., 2020), or LaTeX code (Deng et al., 2019; Kayal et al., 2021). However, this task is still a big challenging problem, due to the diversity of table styles and the complexity of table structures.

In recent years, many table recognition methods have been proposed and proven to be powerful models for both PDF and image-based table recognition. However, most previous studies (Nassar et al., 2022; Qiao et al., 2021; Ye et al., 2021; Zhang et al., 2022) are based on fully supervised learning approaches. They rely on training datasets that contain lots of richly annotated table images. Detailed table image annotation, e.g., cell or text bounding box annotation, however, is costly and often subjective. Due to the rapid development of deep learning, some works (Deng et al., 2019; Zhong et al., 2020) focus on weakly supervised approaches that rely only on HTML (or LaTeX) code-level labels of table images. However, their performance is still mediocre compared to the fully supervised methods.

This paper proposes a weakly supervised model named WSTabNet for image-based table recognition. The proposed model consists of three main parts: an encoder for feature extraction, a structure decoder for generating table structure, and a cell decoder for predicting the content of each cell in the table. The system is trained in an end-to-end manner by stochastic gradient descent algorithms, requiring only

---

[a] https://orcid.org/0000-0002-0856-3196
[b] https://orcid.org/0000-0002-9061-7949
[c] https://orcid.org/0000-0003-1679-723X
[d] https://orcid.org/0000-0002-2909-7163

table images and their ground-truth HTML (or LaTeX) representations. To facilitate table recognition with deep learning, we create and release WikiTableSet[5], the largest publicly available image-based table recognition dataset built from Wikipedia.

The experimental results on WikiTableSet and two large-scale datasets: FinTabNet (Zheng et al., 2021) and PubTabNet (Zhong et al., 2020) show that our model achieves better or similar accuracies compared to the state-of-the-art models on all benchmark datasets. We also evaluated the proposed model on the final evaluation set of the ICDAR2021 competition on scientific literature parsing (Jimeno Yepes et al., 2021), demonstrating that the proposed weakly supervised model outperforms the 5th ranking solution and achieves competitive results when compared to the top four solutions. The code and WikiTableSet dataset will be publicly released to GitHub.

In summary, the main contributions of this paper are as follows:

- We present a novel weakly supervised learning model named WSTabNet for image-based table recognition. The proposed model can be trained end-to-end, requiring only table images and their ground-truth HTML (or LaTeX) representations.
- We present WikiTableSet, the largest publicly available image-based table recognition dataset in three languages built from Wikipedia.
- Across all benchmark datasets, the proposed weakly supervised model achieves better, or similar accuracies compared to the state-of-the-art models.

The rest of this paper is organized as follows. In Sec.2, we give a brief overview of the related works. We introduce the overview of the proposed model in Sec. 3. Sec. 4 describes the overview of the WikiTableSet dataset. In Sec. 5, we report the experimental details and results. Finally, we draw conclusions in Sec. 6.

## 2 RELATED WORK

Table analysis in unstructured documents can be divided into two main parts: table detection (Casado-García et al., 2020; Huang et al., 2019) and table recognition (Deng et al., 2019; Ye et al., 2021; Zhong et al., 2020), which we will briefly survey here. Most of the previous methods for table recognition are based on two-step approaches dividing the problem into two steps. The first step is table structure recognition, which recognizes the table structure (including cell location) from a table image. The second step is cell content recognition that predicts the text content of each table cell from its location. Due to the simplicity of the cell content recognition, which can be easily solved by the standard OCR model (Lu et al., 2021; Ly et al., 2021; Shi et al., 2017), many researchers only focused on the table structure recognition problem (Prasad et al., 2020; Raja et al., 2020a; Schreiber et al., 2017). Early works of table structure recognition are based on hand-crafted features and heuristic rules and mainly applied to simple table structure or pre-defined table structure formats (Itonori, 1993; Kieninger, 1998; Wang et al., 2004). In recent years, motivated by the success of deep learning, especially in object detection and semantic segmentation, many deep learning-based methods (Prasad et al., 2020; Raja et al., 2020a; Schreiber et al., 2017) have been proposed and proven to be powerful models for table structure recognition. S. Schreiber et al. (Schreiber et al., 2017) proposed a two-fold system named DeepDeSRT that applies Faster RCNN (Ren et al., 2015) and FCN (Long et al., 2015) for both table detection and row/column segmentation. Sachin et al. (Raja et al., 2020a) proposed a table structure recognizer named TabStruct-Net that predicts the aligned cell regions and the localized cell relations in a joint manner.

Recently, some researchers (Nassar et al., 2022; Qiao et al., 2021; Ye et al., 2021; Zhang et al., 2022) worked on both table structure recognition and cell content recognition to build a complete table recognition system. J. Ye et al. (Ye et al., 2021) employed the Transformer decoder layers to build the model named TableMASTER for table structure recognition and combined it with a text line detector to detect text lines in each table cell. Finally, they employed a text line recognizer based on (Lu et al., 2021) to recognize each text line in table cells. Their system achieved second place in ICDAR2021 competition (Jimeno Yepes et al., 2021). A. Nassar et al. (Nassar et al., 2022) proposed a Transformer-based model named TableFormer for recognizing both table structure and the bounding box of each table cell and then using the cell contents extracted from the PDF to build the whole table recognition system. Z. Zhang et al. (Zhang et al., 2022) proposed a table structure recognizer named Split, Embed, and Merge (SEM) for recognizing the table structure from a table image. Then, they combined SEM with an attention-based text recognizer to build the table recognizer and achieved third place in the ICDAR2021 competition (Jimeno Yepes et al., 2021). These two-step approaches achieved competitive accuracies; however, these systems rely on training datasets containing many richly annotated table images. They are also difficult to be maintained and inferred due to consisting of multiple separate components.

---
[5] https://github.com/namtuanly/WikiTableSet

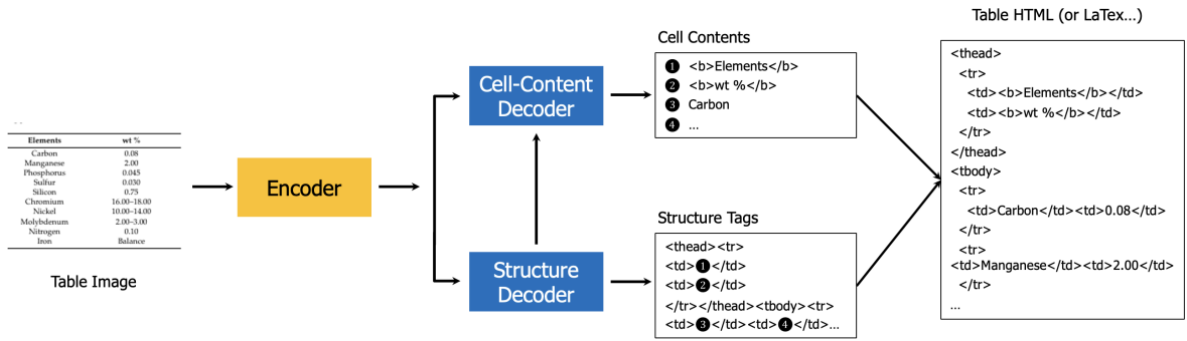

Figure 1. The overview of WSTabNet.

Most recently, due to the progress of Deep Neural Networks especially the encoder-decoder models, some researchers (Deng et al., 2019; Zhong et al., 2020) try to focus on weakly supervised learning approaches that require only HTML (or LaTeX) code level annotations of table images for training the models. Y. Deng et al. (Deng et al., 2019) formulated table recognition as an image-to-LaTeX problem and directly used IM2TEX (Deng et al., 2016) model for recognizing LaTeX representations from a table image. X. Zhong et al. (Zhong et al., 2020) proposed an encoder-dual-decoder (EDD) model for recognizing table structure and cell content. They also publicized a table recognition dataset (PubTabNet) to the community. These methods significantly reduce the annotation cost of the training data; however, their performance is still mediocre compared to the fully supervised learning methods.

In this work, we propose a weakly supervised learning model that reduces the annotation cost and achieves competitive performance compared to the fully supervised learning methods.

## 3 THE PROPOSED METHOD

WSTabNet consists of three main components, as shown in Fig. 1: 1) an encoder for feature extraction, 2) a structure decoder for recognizing the structure of the table image, and 3) a cell decoder for predicting the content of each cell in the table. The encoder extracts the features from the input table image and encodes them as a sequence of features. The sequence of features is passed into the structure decoder to predict a sequence of HTML structure tags that represents the table's structure. When the structure decoder produces the HTML structure tag representing a new cell ('<td>' or '<td …>'), the hidden output of the structure decoder corresponding to that cell is passed into the cell decoder to predict the text content of that cell. Finally, the text contents of cells are inserted into the HTML structure tags corresponding to their cells to produce the final HTML code representing the input table image. Fig. 2 shows the detail of the three components in our model. We describe the detail of each component in the following sections.

### 3.1 Encoder

In this work, we employ a CNN backbone network followed by a positional encoding layer to build the encoder.

Given an input table image, the CNN backbone extracts a feature grid $F$ of the size $\{h', w', k\}$ (where $k$ is the number of the feature maps, $h'$ and $w'$ depend on the input image size and the number of the pooling layers). The feature grid $F$ will be unfolded into a sequence of features (column by column from left to right in each feature map) and then fed into the positional encoding layer to get the encoded sequence of features. Finally, the encoded sequence of features is fed into the two decoders to predict the table structure and the contents of the cells.

### 3.2 Structure Decoder

At the top of the encoder, the structure decoder uses the outputs of the encoder to predict a sequence of HTML tags representing the table structure. Inspired by the works of (Ye et al., 2021; Zhong et al., 2020), the HTML tags of the table structure are tokenized at the HTML tag level except for the tag of a cell. In our model, the form of '<td></td>' is treated as one structure token class. Note that this can largely reduce the length of the sequence. We also break down the tag of the spanning cells ('<td rowspan/colspan="*number*">') into '<td', 'rowspan=' or 'colspan=', with the number of spanning cells, and '>'. Thus, the structure token of '<td></td>' or '<td' represents a new table cell.

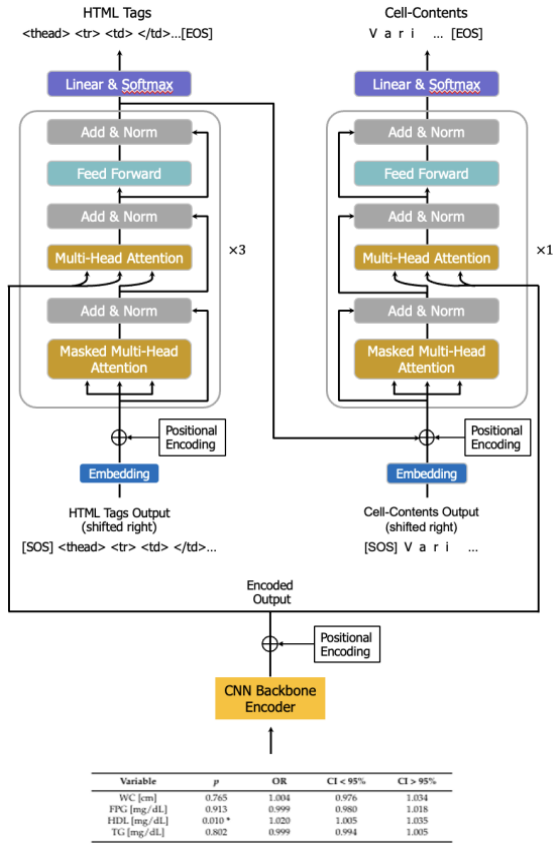

Figure 2. Network architecture of WSTabNet.

The architecture of this component is inspired by the Transformer decoder (Vaswani et al., 2017), which consists of a stack of *N* identical Transformer decoder layers (identical layer in short) that comprise mainly of multi-head attention and feed-forward layers. As shown in Fig. 2, the structure decoder comprises a stack of three identical layers followed by a linear layer and a softmax layer. The bottom identical layer takes the outputs of the encoder as the key and value vector inputs. During training, the right-shifted sequence of target HTML tags (structure tokens) of the table structure (after passing through the embedded layer and the positional encoding layer) is passed into the bottom identical layer as the query vector. In the inference stage, the right-shifted sequence of target HTML tags is replaced by the right-shifted sequence of HTML tags outputted by the structure decoder. Finally, the output of the top identical layer is fed into the linear layer. Then, the results are fed into the softmax layer to generate the sequence of structure tokens representing the table structure.

### 3.3 Cell Decoder

When the structure decoder produces the structure token representing a new cell ('<td>' or '<td …>'), the hidden output of the structure decoder corresponding to that structure token and the output of the encoder are passed into the cell decoder to predict the content of the cell. The cell decoder in our model can be considered a text recognizer. The text contents of cells are tokenized at the character level.

In this work, we use one identical layer followed by a linear layer and a softmax layer to build the cell decoder, as shown in Fig. 2. The identical layer takes the output of the encoder as the input value and key vectors. During training, the right-shifted target text content of the cell is passed through the embedded and the positional encoding layers and then added to the hidden output of the structure decoder before being fed into the identical layer as the query vector. In the inference state, the right-shifted target text content of the cell is replaced by the right-shifted text content of the cell outputted by the cell decoder. Finally, the output of the identical layer is fed into the linear layer and then the softmax layer to generate the contents of the cells.

### 3.4 Network Training

The whole system can be trained in an end-to-end manner by stochastic gradient descent algorithms on pairs of a table image and its annotation of HTML code. The overall loss of our model is defined as follows:

$$\mathcal{L} = \lambda \mathcal{L}_{\text{struc.}} + (1 - \lambda) \mathcal{L}_{\text{cell}} \quad (1)$$

where $\mathcal{L}_{\text{struc.}}$ and $\mathcal{L}_{\text{cell}}$ are the cross-entropy loss of generating the structure tokens and predicting the cell tokens, respectively. $\lambda \in [0,1]$ is a weight hyperparameter.

## 4 WIKITABLESET

Many datasets for table (structure) recognition have recently been released to the research community

(Zhong et al., 2020). However, there are few large datasets for table recognition that have enough data to support the training of deep learning-based methods, including PubTabNet (Zhong et al., 2020), FinTabNet (Zheng et al., 2021), and PubTables-1M (Smock Brandon et al., 2022). We summarize these datasets in Table 1. Furthermore, as far as we know, all the current datasets for table recognition are table datasets in English, and currently, we are missing the dataset in other languages.

To facilitate the research of table recognition in different languages with deep learning, we release WikiTableSet, the largest publicly available image-based table recognition dataset built from Wikipedia. WikiTableSet contains nearly 4 million English table images, 590K Japanese table images, 640k French table images with corresponding HTML representation, and cell bounding boxes. We build a Wikipedia table extractor[6] and use this to extract tables (in HTML code format) from the 2022-03-01 dump of Wikipedia. In this study, we select Wikipedia tables from three representative languages, i.e., English, Japanese, and French; however, the dataset could be extended to around 300 languages with 17M tables using our table extractor. Second, we normalize the HTML tables following the PubTabNet format (separating table headers and table data; removing CSS and style tags). Finally, we use Chrome and Selenium to render table images from table HTML codes. This dataset provides a standard benchmark for studying table recognition algorithms in different languages or even multilingual table recognition algorithms.

Table 1: Comparison of table datasets for table structure recognition (TSR) and table recognition (TR).

| Datasets | Samples | TSR | TR | Language |
|---|---|---|---|---|
| PubTabNet | 510K | ✓ | ✓ | English |
| FinTabNet | 113K | ✓ | ✓ | English |
| PubTables-1M | 948K | ✓ | ✓ | English |
| WikiTableSet (ours) | 5M | ✓ | ✓ | English Japanese French |

## 5 EXPERIMENTS

To evaluate the effectiveness of the proposed model, abundant comparative experiments are carried out on four datasets: FinTabNet (Zheng et al., 2021),

---

[6] https://github.com/phucty/wtabhtml

PubTabNet (Zhong et al., 2020), WikiTableSet_EN750K, and WikiTableSet_JA. The information on the datasets is given in Sec 5.1. The implementation details are described in Sect. 5.2; the experimental results are presented in Sect. 5.3; and the visualization results are shown in Sect. 5.4.

### 5.1 Datasets

**WikiTableSet-EN750K.** Considering the computational cost for training various deep learning models, we randomly choose 750K English table images from the English part of WikiTableSet to create a subset of English table images which is named WikiTableSet-EN750K. We split the subset into training, testing, and validation sets with a ratio of 9:1:1. Consequently, this subset consists of 619,814 table images for training, 17,014 table images for validation, and 16,842 table images for testing.

**WikiTableSet_JA.** We named a subset of Japanese table images in WikiTableSet as WikiTableSet_JA and divided them into training, testing, and validation sets with a ratio of 9:1:1.

**PubTabNet** (Zhong et al., 2020) is a large-scale table image dataset that consists of over 568k table images with their corresponding annotations of the table HTML and bounding box coordinates of each non-empty table cell. The dataset is divided into 500,777 training samples and 9,115 validation samples used in the development phase, and 9,064 final evaluation samples used in the Final Evaluation Phase of the ICDAR2021 competition (Jimeno Yepes et al., 2021).

**FinTabNet** is another large-scale table image dataset published by X. Zheng et al. (Zheng et al., 2021). The dataset comprises 112k table images from the annual reports of the S&P 500 companies with detailed annotations of the table HTML and cell bounding boxes like PubTabNet. This dataset is divided into training, testing, and validation sets with a ratio of 81% : 9.5% : 9.5%.

### 5.2 Implementation Details

In the encoder, we use the ResNet-31 network (He et al., 2016) with the Multi-Aspect Global Context Attention (GCAttention) (Lu et al., 2021) after each residual block to build the CNN backbone. All table images are resized to 480x480 before feeding into the CNN backbone, and the feature map outputted from the CNN backbone will be had a dimension of 60x60.

All identical layers in the two decoders have the same architecture with 8 attention heads, the input feature size, and the feed-forward network size are 512 and 2048, respectively. The maximum length of a sequence of structure tokens and a sequence of cell tokens are 500 and 150, respectively. The weight hyperparameter is set as $\lambda = 0.5$.

WSTabNet is implemented with PyTorch and the MMCV library (MMCV Contributors, 2018). The model is trained on two NVIDIA A100 80G with a batch size of 8. The initializing learning rate is 0.001 for the first 12 epochs. Afterward, we reduce the learning to 0.0001 and train for 5 more epochs or convergence.

## 5.3 Experimental Results

To evaluate the performance of the proposed model, we employ the Tree-Edit-Distance-Based Similarity (TEDS) metric (Zhong et al., 2020). It represents the prediction and the ground-truth as a tree structure of HTML tags, and is calculated as follows:

$$\text{TEDS}(T_a, T_b) = 1 - \frac{\text{EditDist}(T_a, T_b)}{\max(|T_a|, |T_b|)} \quad (2)$$

where $T_a$ and $T_b$ represent tables in a tree structured HTML format, EditDist denotes the tree-edit distance, and $|T|$ represents the number of nodes in T.

We denote TEDS-struc. as the TEDS score between two tables when considering only the table structure information.

### 5.3.1 Table Recognition

In the first experiment, we evaluated the performance of the proposed model for table recognition on FinTabNet, PubTabNet, WikiTableSet_EN750K and WikiTableSet_JA. Table 2 shows the table recognition performance (TEDS) of the proposed model on all benchmark datasets, and Table 3 shows the comparison between its performance with the previous methods on PubTabNet.

As shown in Table 2, the proposed model achieved TEDS of more than 92.5% on both simple and complex table images of all benchmark datasets. The results imply that the proposed model works well on both simple and complex table images as well as both English and Japanese table images. As shown in Table 3, on the PubTabNet dataset, the proposed model achieves superior performance compared to the previous methods. Specifically, with TEDS of 96.48%, our model significantly improves the weakly supervised learning methods of EDD and IM2TEX by more than 8%. Although, the proposed model requires only table HTML annotations for the training step, it outperforms all the fully supervised methods (Nassar et al., 2022; Qiao et al., 2021; Raja et al., 2020b; Ye et al., 2021; Zhang et al., 2022; Zheng et al., 2021) that require both table HTML and the cell bounding boxes annotations for training the models. Note that all the previous fully supervised methods are non-end-to-end approach. VCGoup's solution, and SEM are the 2nd ranking, and 3rd ranking solutions in ICDAR2021 competition, respectively. LGPMA (Qiao et al., 2021) is the table structure recognizer component in the 1st ranking solution in ICDAR2021 competition.

Table 2: Table recognition results on PubTabNet (PTN), FinTabNet (FTN), WikiTableSet_EN750K (WTN[EN]) and WikiTableSet_JA (WTN[JA]).

| Model | TEDS (%) | | | |
|---|---|---|---|---|
| | Dataset | S | C | All |
| WSTabNet | PTN | **97.89** | **95.02** | **96.48** |
| WSTabNet | FTN | 95.24 | 95.41 | 95.32 |
| WSTabNet | WTS[EN] | 95.37 | 92.87 | 94.83 |
| WSTabNet | WTS[JA] | 93.82 | 92.56 | 93.43 |

S (Simple): Tables without multi-column or multi-row cells.
C (Complex): Tables with multi-column or multi-row cells.

Table 3: The comparison between WSTabNet and the previous methods on PubTabNet (PTN).

| Model | TEDS (%) | | |
|---|---|---|---|
| | Simp. | Comp. | All |
| IM2TEX (Deng et al., 2019) | 81.70 | 75.50 | 78.60 |
| EDD (Zhong et al., 2020) | 91.20 | 85.40 | 88.30 |
| TabStruct-Net (Raja et al., 2020b) | - | - | 90.10 |
| GTE (Zheng et al., 2021) | - | - | 93.00 |
| TableFormer (Nassar et al., 2022) | 95.40 | 90.10 | 93.60 |
| SEM [3] (Zhang et al., 2022) | 94.80 | 92.50 | 93.70 |
| LGPMA [1] (Qiao et al., 2021) | - | - | 94.60 |
| VCGoup [2] (Ye et al., 2021) | - | - | 96.26 |
| **WSTabNet** | **97.89** | **95.02** | **96.48** |

Simp. (Simple): Tables without multi-column or multi-row cells.
Comp. (Complex): Tables with multi-column or multi-row cells.
(1)(2)(3) are 1st, 2nd, and 3rd solutions in ICDAR2021 competition.

We also evaluate the proposed model on the final evaluation set of PubTabNet which is used for the

Final Evaluation Phase in ICDAR2021 competition. Table 4 compares TEDS scores by the proposed model and the top 10 solutions in ICDAR2021 competition (Jimeno Yepes et al., 2021). Although we used only table HTML information for training and neither any additional training data nor ensemble techniques, the proposed model outperforms the 5th ranking solution named DBJ and achieves competitive results when compared to the 4th ranking solution in the final evaluation set of Task-B in the ICDAR 2021 competition. Note that all top 10 solutions are non-end-to-end approaches and require both table HTML and the cell bounding boxes information for the training step. Furthermore, most of them also use additional data for training as well as ensemble methods.

Table 4: Table recognition results on PubTabNet final evaluation set.

| Team Name | TEDS | | |
|---|---|---|---|
| | Simp. | Comp. | All |
| Davar-Lab-OCR | 97.88 | 94.78 | **96.36** |
| VCGroup (Ye et al., 2021) | **97.90** | 94.68 | 96.32 |
| XM (Zhang et al., 2022) | 97.60 | **94.89** | 96.27 |
| YG | 97.38 | 94.79 | 96.11 |
| **WSTabNet (Our)** | 97.51 | 94.37 | 95.97 |
| DBJ | 97.39 | 93.87 | 95.66 |
| TAL | 97.30 | 93.93 | 95.65 |
| PaodingAI | 97.35 | 93.79 | 95.61 |
| anyone | 96.95 | 93.43 | 95.23 |
| LTIAYN | 97.18 | 92.40 | 94.84 |

Simp. (Simple): Tables without multi-column or multi-row cells.
Comp. (Complex): Tables with multi-column or multi-row cells.

### 5.3.2 Table Structure Recognition

In this experiment, we evaluate the effectiveness of WSTabNet for recognizing the structure of the table images on PubTabNet and FinTabNet datasets. Table 5 shows the table structure recognition performance (TEDS-struc. scores) of the proposed model and the previous table structure recognition methods on the validation set of the PubTabNet and the test set of the FinTabNet.

On the test set of the FinTabNet dataset, the proposed model achieved TEDS-struc. of 99.06%, 98.33%, and 98.72% on the simple, complex, and all table images, respectively. These results show that the proposed model works well on both simple and complex tables and outperforms EDD (Zhong et al., 2020) by about 8.1% and the best model in (Nassar et al., 2022) by about 2%. On the validation set of the PubTabNet dataset, the proposed model achieved TEDS-struc. of 97.74% on all table images which again improves EDD by about 7.8% and the best model in (Nassar et al., 2022) by about 1%. Note that all other methods except EDD are fully supervised approaches that require both HTML and cell bounding boxes annotations in the training step.

Table 5: Table structure recognition results on PubTabNet validation set (PTN) and FinTabNet (FTN).

| Dataset | Model | TEDS-struc. (%) | | |
|---|---|---|---|---|
| | | Simp. | Comp. | All |
| FTN | EDD (Zhong et al., 2020) | 88.40 | 92.08 | 90.60 |
| | GTE (Zheng et al., 2021) | - | - | 87.14 |
| | GTE [(FT)] (Zheng et al., 2021) | - | - | 91.02 |
| | TableFormer (Nassar et al., 2022) | 97.50 | 96.00 | 96.80 |
| | **WSTabNet** | **99.06** | **98.33** | **98.72** |
| PTN | EDD (Zhong et al., 2020) | 91.10 | 88.70 | 89.90 |
| | GTE (Zheng et al., 2021) | - | - | 93.01 |
| | LGPMA (Qiao et al., 2021) | - | - | 96.70 |
| | TableFormer (Nassar et al., 2022) | 98.50 | 95.00 | 96.75 |
| | **WSTabNet** | **99.06** | **96.37** | **97.74** |

Simp. (Simple): Tables without multi-column or multi-row cells.
Comp. (Complex): Tables with multi-column or multi-row cells.
(FT) Model was trained on PubTabNet and then finetuned.

## 5.4 Visualization Results

In this section, we show some visualization results of WSTabNet on PubTabNet. As shown in Fig. 3, the left image is the original table image, and the right image is the presentation on the web browser of the generated HTML code. As it is shown, WSTabNet can recognize the complex table structure and the cell contents, even for the empty cells or cells that span multiple rows/columns.

## 6 CONCLUSIONS

In this paper, we present a weakly supervised learning model named WSTabNet for image-based table recognition that relies only on HTML (or LaTeX) code level labels of table images. The proposed model consists of three main parts: an encoder for feature extraction, a structure decoder for generating table structure, and a cell decoder for predicting the content of each cell in the table. The whole system is trained end-to-end by stochastic gradient descent algorithms. To facilitate the research of table recognition in different languages with deep learning, we create and release WikiTableSet, the largest publicly image-based table recognition dataset in three languages built from Wikipedia. Extensive experiments on WikiTableSet and two large-scale datasets demonstrate that the proposed model achieves competitive accuracies compared to the fully supervised learning methods.

Figure 4. Visualization results on PubTabNet.

In the future, we will expand the WikiTableSet to more languages and conduct experiments of the proposed model on the table image datasets of different languages.

## ACKNOWLEDGEMENTS

This work was supported by the Cross-ministerial Strategic Innovation Promotion Program (SIP) Second Phase and "Big-data and AI-enabled Cyberspace Technologies" by New Energy and Industrial Technology Development Organization (NEDO).

## REFERENCES

Casado-García, Á., Domínguez, C., Heras, J., Mata, E., & Pascual, V. (2020). The benefits of close-domain fine-tuning for table detection in document images. *Lecture Notes in Computer Science (Including Subseries Lecture Notes in Artificial Intelligence and Lecture Notes in Bioinformatics)*, *12116 LNCS*, 199–215. https://doi.org/10.1007/978-3-030-57058-3_15

Deng, Y., Kanervisto, A., Ling, J., & Rush, A. M. (2016). Image-to-Markup Generation with Coarse-to-Fine Attention. *34th International Conference on Machine Learning, ICML 2017*, *3*, 1631–1640. https://doi.org/10.48550/arxiv.1609.04938

Deng, Y., Rosenberg, D., & Mann, G. (2019). Challenges in end-to-end neural scientific table recognition. *Proceedings of the International Conference on Document Analysis and Recognition, ICDAR*, 894–901. https://doi.org/10.1109/ICDAR.2019.00148

He, K., Zhang, X., Ren, S., & Sun, J. (2016). Deep Residual Learning for Image Recognition. *2016 IEEE Conference on Computer Vision and Pattern Recognition (CVPR)*, 770–778. https://doi.org/10.1109/CVPR.2016.90

Huang, Y., Yan, Q., Li, Y., Chen, Y., Wang, X., Gao, L., & Tang, Z. (2019). A YOLO-based table detection method. *Proceedings of the International Conference on Document Analysis and Recognition, ICDAR*, 813–818. https://doi.org/10.1109/ICDAR.2019.00135

Itonori, K. (1993). Table structure recognition based on textblock arrangement and ruled line position. *Proceedings of 2nd International Conference on Document Analysis and Recognition (ICDAR '93)*, 765,766,767,768-765,766,767,768. https://doi.org/10.1109/ICDAR.1993.395625

Jimeno Yepes, A., Zhong, P., & Burdick, D. (2021). ICDAR 2021 Competition on Scientific Literature Parsing. *Lecture Notes in Computer Science (Including Subseries Lecture Notes in Artificial Intelligence and Lecture Notes in Bioinformatics)*, *12824 LNCS*, 605–617. https://doi.org/10.1007/978-3-030-86337-1_40

Kayal, P., Anand, M., Desai, H., & Singh, M. (2021). ICDAR 2021 Competition on Scientific Table Image Recognition to LaTeX. *Lecture Notes in Computer Science (Including Subseries Lecture Notes in Artificial Intelligence and Lecture Notes in Bioinformatics)*, *12824 LNCS*, 754–766. https://doi.org/10.1007/978-3-030-86337-1_50

Kieninger, T. G. (1998). Table structure recognition based on robust block segmentation. *Https://Doi.Org/10.1117/12.304642*, *3305*, 22–32. https://doi.org/10.1117/12.304642

Li, M., Cui, L., Huang, S., Wei, F., Zhou, M., & Li, Z. (2019). *TableBank: A Benchmark Dataset for Table Detection and Recognition*. https://doi.org/10.48550/arxiv.1903.01949

Long, J., Shelhamer, E., & Darrell, T. (2015). Fully convolutional networks for semantic segmentation. *2015 IEEE Conference on Computer Vision and Pattern Recognition (CVPR)*, *07-12-June-2015*, 3431–3440. https://doi.org/10.1109/CVPR.2015.7298965

Lu, N., Yu, W., Qi, X., Chen, Y., Gong, P., Xiao, R., & Bai, X. (2021). MASTER: Multi-aspect non-local network for scene text recognition. *Pattern Recognition*, *117*, 107980. https://doi.org/10.1016/J.PATCOG.2021.107980

Ly, N. T., Nguyen, H. T., & Nakagawa, M. (2021). 2D Self-attention Convolutional Recurrent Network for Offline Handwritten Text Recognition. *Lecture Notes in Computer Science (Including Subseries Lecture Notes in Artificial Intelligence and Lecture Notes in Bioinformatics)*, *12821 LNCS*, 191–204. https://doi.org/10.1007/978-3-030-86549-8_13/COVER

MMCV Contributors. (2018). *{MMCV: OpenMMLab} Computer Vision Foundation*. https://github.com/open-mmlab/mmcv

Nassar, A., Livathinos, N., Lysak, M., & Staar, P. (2022). *TableFormer: Table Structure Understanding with Transformers*. https://doi.org/10.48550/arxiv.2203.01017

Prasad, D., Gadpal, A., Kapadni, K., Visave, M., & Sultanpure, K. (2020). CascadeTabNet: An approach for end to end table detection and structure recognition from image-based documents. *IEEE Computer Society Conference on Computer Vision and Pattern Recognition Workshops*, *2020-June*, 2439–2447. https://doi.org/10.48550/arxiv.2004.12629

Qiao, L., Li, Z., Cheng, Z., Zhang, P., Pu, S., Niu, Y., Ren, W., Tan, W., & Wu, F. (2021). LGPMA:

Complicated Table Structure Recognition with Local and Global Pyramid Mask Alignment. *Lecture Notes in Computer Science (Including Subseries Lecture Notes in Artificial Intelligence and Lecture Notes in Bioinformatics)*, *12821 LNCS*, 99–114. https://doi.org/10.1007/978-3-030-86549-8_7/TABLES/4

Raja, S., Mondal, A., & Jawahar, C. v. (2020a). Table Structure Recognition Using Top-Down and Bottom-Up Cues. *Lecture Notes in Computer Science (Including Subseries Lecture Notes in Artificial Intelligence and Lecture Notes in Bioinformatics)*, *12373 LNCS*, 70–86. https://doi.org/10.1007/978-3-030-58604-1_5/FIGURES/8

Raja, S., Mondal, A., & Jawahar, C. v. (2020b). Table Structure Recognition Using Top-Down and Bottom-Up Cues. *Lecture Notes in Computer Science (Including Subseries Lecture Notes in Artificial Intelligence and Lecture Notes in Bioinformatics)*, *12373 LNCS*, 70–86. https://doi.org/10.1007/978-3-030-58604-1_5

Ren, S., He, K., Girshick, R., & Sun, J. (2015). Faster R-CNN: Towards Real-Time Object Detection with Region Proposal Networks. *IEEE Transactions on Pattern Analysis and Machine Intelligence*, *39*(6), 1137–1149. https://doi.org/10.48550/arxiv.1506.01497

Schreiber, S., Agne, S., Wolf, I., Dengel, A., & Ahmed, S. (2017). DeepDeSRT: Deep Learning for Detection and Structure Recognition of Tables in Document Images. *Proceedings of the International Conference on Document Analysis and Recognition, ICDAR*, *1*, 1162–1167. https://doi.org/10.1109/ICDAR.2017.192

Shi, B., Bai, X., & Yao, C. (2017). An End-to-End Trainable Neural Network for Image-Based Sequence Recognition and Its Application to Scene Text Recognition. *IEEE Transactions on Pattern Analysis and Machine Intelligence*, *39*(11), 2298–2304. https://doi.org/10.1109/TPAMI.2016.2646371

Smock Brandon, Pesala Rohith, & Abraham Robin. (2022). PubTables-1M: Towards comprehensive table extraction from unstructured documents. *Proceedings of the IEEE/CVF Conference on Computer Vision and Pattern Recognition (CVPR)*, 4634–4642.

Vaswani, A., Shazeer, N., Parmar, N., Uszkoreit, J., Jones, L., Gomez, A. N., Kaiser, Ł., & Polosukhin, I. (2017). Attention is all you need. *Advances in Neural Information Processing Systems*, *2017-December*, 5999–6009.

Wang, Y., Phillips, I. T., & Haralick, R. M. (2004). Table structure understanding and its performance evaluation. *Pattern Recognition*, *37*(7), 1479–1497. https://doi.org/10.1016/J.PATCOG.2004.01.012

Ye, J., Qi, X., He, Y., Chen, Y., Gu, D., Gao, P., & Xiao, R. (2021). *PingAn-VCGroup's Solution for ICDAR 2021 Competition on Scientific Literature Parsing Task B: Table Recognition to HTML*. https://doi.org/10.48550/arxiv.2105.01848

Zhang, Z., Zhang, J., Du, J., & Wang, F. (2022). Split, Embed and Merge: An accurate table structure recognizer. *Pattern Recognition*, *126*, 108565. https://doi.org/10.1016/J.PATCOG.2022.108565

Zheng, X., Burdick, D., Popa, L., Zhong, X., & Wang, N. X. R. (2021). Global Table Extractor (GTE): A Framework for Joint Table Identification and Cell Structure Recognition Using Visual Context. *2021 IEEE Winter Conference on Applications of Computer Vision (WACV)*, 697–706. https://doi.org/10.1109/WACV48630.2021.00074

Zhong, X., ShafieiBavani, E., & Jimeno Yepes, A. (2020). Image-Based Table Recognition: Data, Model, and Evaluation. *Lecture Notes in Computer Science (Including Subseries Lecture Notes in Artificial Intelligence and Lecture Notes in Bioinformatics)*, *12366 LNCS*, 564–580. https://doi.org/10.1007/978-3-030-58589-1_34/TABLES/3